\documentclass[12pt]{article}
\usepackage{multirow}
\usepackage{booktabs}
\usepackage{mathrsfs}
\usepackage{fullpage}
\usepackage{amsmath}
\usepackage{graphicx}%
\usepackage{amsfonts}%
\usepackage{amssymb}
\usepackage{afterpage}
\usepackage{xurl}
\usepackage{fancyhdr}
\usepackage{bm}
\usepackage{isomath}
\usepackage{amsmath}
\usepackage{amsbsy}
\usepackage{amssymb}
\usepackage{amscd}
\usepackage{amsfonts}
\usepackage{graphicx}

\usepackage{verbatim}
\usepackage{euscript}
\usepackage{alltt}
\usepackage{stmaryrd}
\usepackage{float}
\usepackage[titletoc,title]{appendix}
\usepackage{caption}
\usepackage[english]{babel}
\usepackage{enumitem}
\usepackage{layout}
\usepackage{longtable}
\usepackage{tikz}
\usepackage{amsmath}
\usepackage{bm}
\usepackage{amsfonts}

\newcommand{\bfsigma}{\bm{\sigma}}
\newcommand{\bfnabla}{\bm{\nabla}}
\newcommand{\bbC}{\mathbb{C}}
\newcommand{\bfepsilon}{\bm{\epsilon}}
\newcommand{\bft}{\bm{t}}
\newcommand{\bfu}{\bm{u}}
\newcommand{\bfn}{\bm{n}}
\newcommand{\bfB}{\bm{B}}
\usepackage{subfigure}

\usepackage[linktoc=all]{hyperref}
\hypersetup{colorlinks=true,
	linkcolor=blue,
	citecolor=blue
}

\usepackage[margin=1in,  headheight=3ex, headsep=2ex]{geometry}
\usepackage{titlesec}

\titleformat{\paragraph}
{\normalfont\normalsize\bfseries}{\theparagraph}{1em}{}
\titlespacing*{\paragraph}
{0pt}{3.25ex plus 1ex minus .2ex}{1.5ex plus .2ex}

\PassOptionsToPackage{normalem}{ulem}
\usepackage{ulem}

\newcommand\blfootnote[1]{%
  \begingroup
  \renewcommand\thefootnote{}\footnote{#1}%
  \addtocounter{footnote}{-1}%
  \endgroup
}

\begin{document}

%\pagenumbering{roman}

\title{Machine learning-accelerated computational solid mechanics: Application to  linear elasticity}

\author{Rajat Arora\thanks{Senior Member of Technical Staff at Advanced Micro Devices, Inc (AMD). rajat.arora9464@gmail.com.}}

\date{}% not using this gives today's date

\maketitle

\begin{abstract}
	This work presents a novel physics-informed deep learning based super-resolution framework to reconstruct high-resolution deformation fields from low-resolution counterparts, obtained from coarse mesh simulations or experiments. We leverage the governing equations and boundary conditions of the physical system to train the model without using any high-resolution labeled data. The proposed approach is applied to obtain the super-resolved deformation fields from the low-resolution stress and displacement fields obtained by running simulations on a coarse mesh for a body undergoing linear elastic deformation. We  demonstrate that the super-resolved fields match the accuracy of an advanced numerical solver running at $400$ times the coarse mesh resolution, while simultaneously satisfying the governing laws. A brief evaluation study comparing the performance of two deep learning based super-resolution architectures  is also presented. 
\blfootnote{Accepted at AAAI 2022: Workshop on AI to Accelerate Science and Engineering (AI2ASE)}
%	We also present a brief evaluation study comparing the performance of two deep learning architectures for the super-resolution example discussed. 
	
	% The low-resolution data is obtained by running simulations on on a We  demonstrate that the super-resolved fields match the accuracy of a numerical solver running at $400$ times the coarse mesh resolution 
\end{abstract}

\section{Introduction}
\label{sec:introduction}
\begin{figure}[t]
	\centering
	{\includegraphics[width=.98\linewidth]{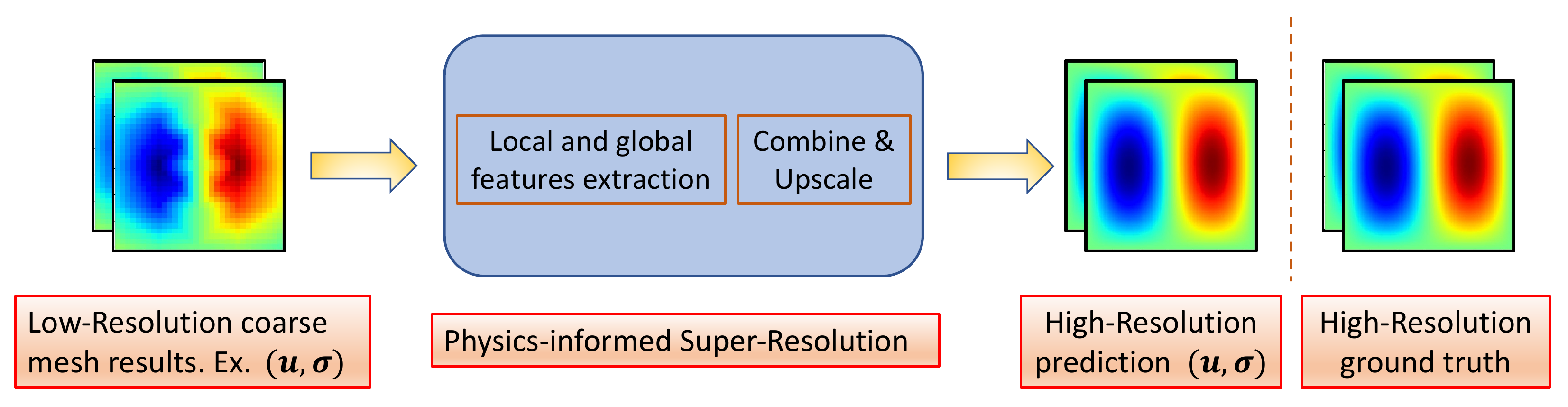}}
	\caption{Schematic of the super-resolution framework.}
	\label{fig:sch_sr}
\end{figure}

% and i in a body undergoing elastic deformation
Image super-resolution (SR) is an active area of research in the field of computer science which aims at recovering high-resolution (HR) image from a low-resolution (LR) image. In this work, we  focus on exploring the concept of image super-resolution to develop a physics-informed Deep Learning (DL) model to reconstruct HR deformation fields (stress and  displacements) from LR fields without requiring any HR labeled data. The LR data could be obtained by running simulations on a coarse mesh or from experiments such as digital image correlation.  We also present a brief study that compares two DL  architectures and evaluate their suitability for developing physics-informed super-resolution framework.   The overall schematic of the proposed physics-informed strategy for super resolution is depicted in Fig.\,\ref{fig:sch_sr}. The use of such physics-informed SR framework will allow researchers to solve computationally expensive simulations much faster and enable them to increase accuracy without additional costs.

Recently, several researchers have explored the possibility of using deep learning based super-resolution to reconstruct HR fluid flow fields from LR (possibly noisy) data. The data-driven approaches for reconstructing HR flow field  \cite{fukami2021machine,fukami2019super,deng2019super,bode2019using,xie2018tempogan} relies on the availability of large amount of HR labeled data. Moreover, the HR output obtained from data-driven approaches may fail to satisfy physics-based constraints because of the lack of any embedded physical constraints  in the model itself. Several studies have demonstrated the merits of developing physics-informed DL models for SR in the fluid mechanics community \cite{esmaeilzadeh2020meshfreeflownet,subramaniam2020turbulence,sun2020physics,gao2021super}. However, to the best of author's knowledge, developing an effective physics-informed
DL model for super-resolution  in label-scarce or label-free scenarios for solid mechanics problems has not yet been explored.

The layout of the rest of this paper is as follows: In Sec.~\ref{sec:theory}, a brief review of the governing equations for modeling elastic deformation in solids is presented. Model architectures and construction of loss function are discussed in Secs.~\ref{sec:arch} and \ref{sec:loss}, respectively. Sec \ref{sec:results} presents the findings for the evaluation of the proposed SR-framework after briefly discussing simulation setup and data collection strategy. Conclusions and future opportunities are presented in Sec.~\ref{sec:conclusion}.

\begin{figure*}[htp]
	\centering
	\subfigure[]{\includegraphics[width=.20\linewidth]{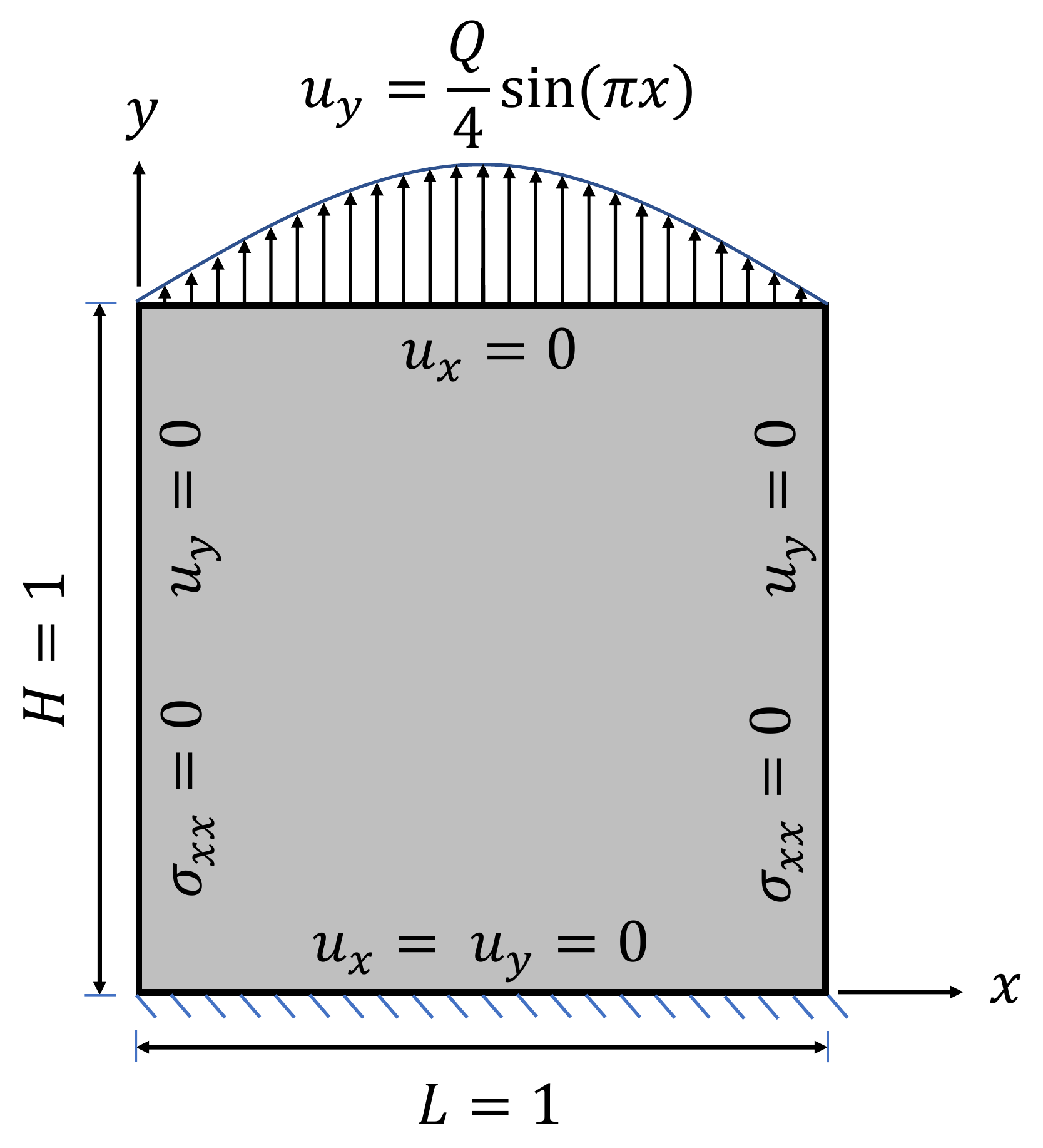}}\label{fig:sch_body}\qquad
	\subfigure[]{\includegraphics[width=.15\linewidth]{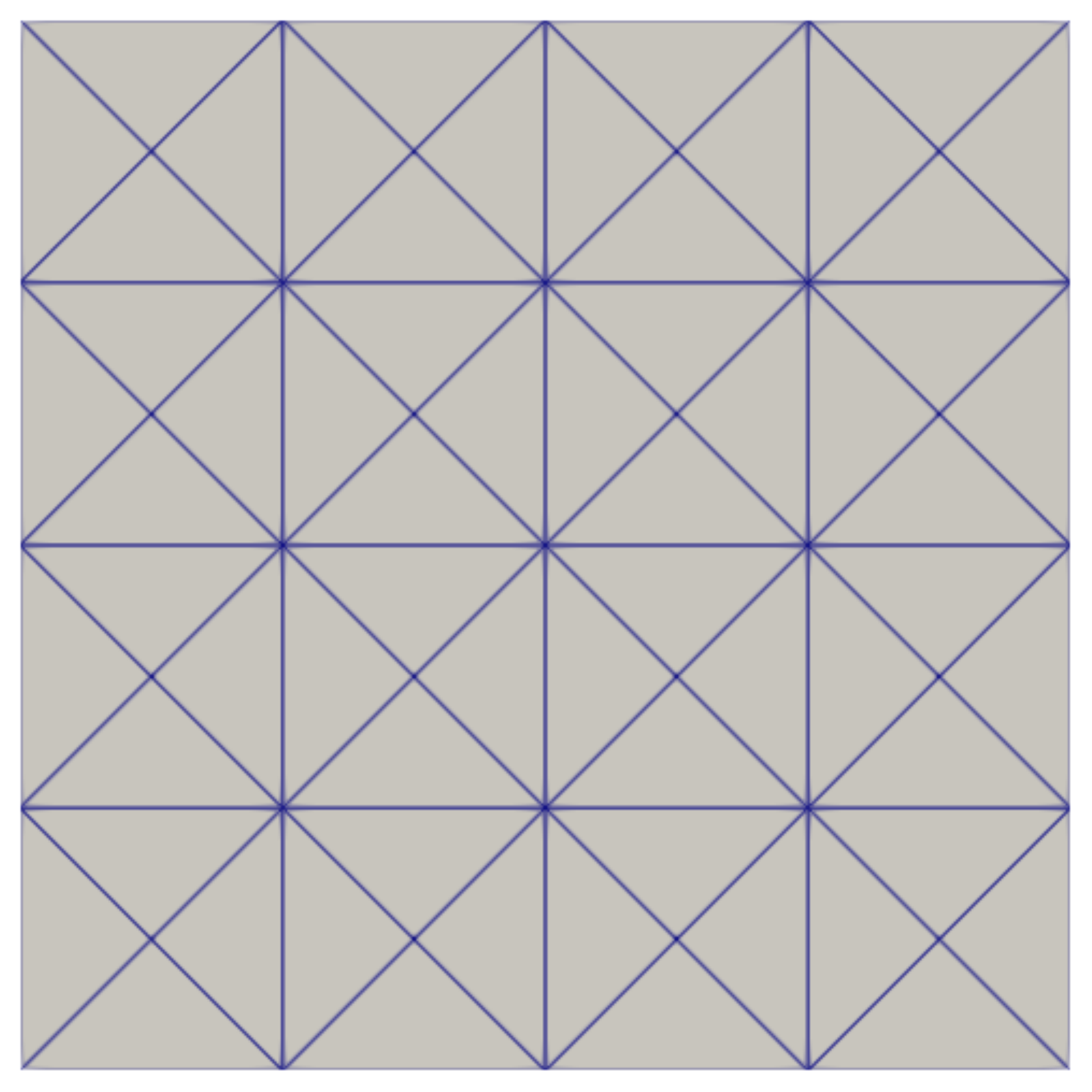}}\label{fig:coarse_mesh}\qquad
	\subfigure[]{\includegraphics[width=.15\linewidth]{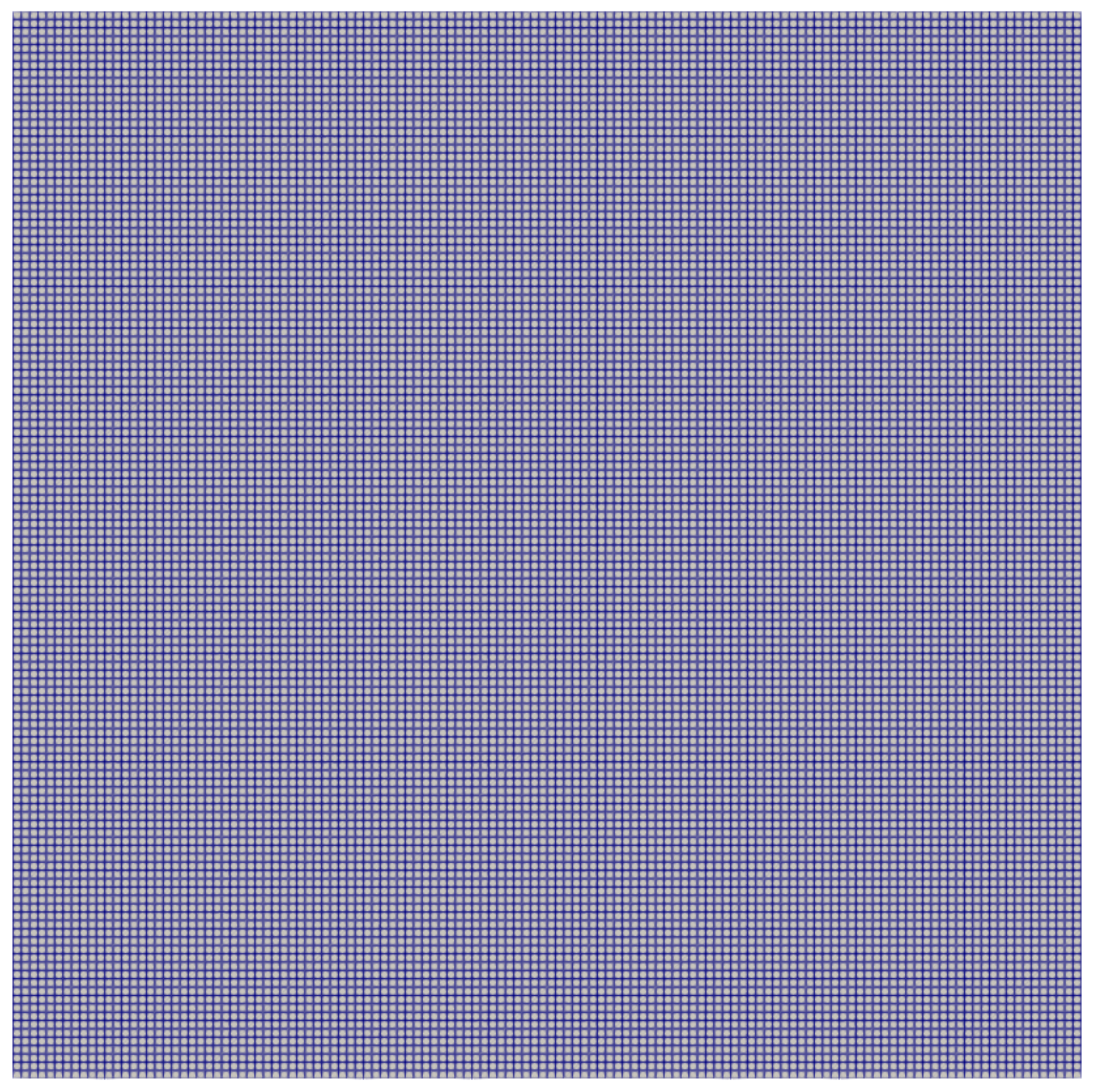}}\label{fig:fine_mesh}
	\caption{a) Schematic showing the geometry and the applied boundary conditions. b) Coarse triangular mesh with $41$ nodes. c) $128 \times 128$ fine mesh with $16384$ nodes. The LR data is refined by $400$ times.}
	\label{fig:fig_army}
\end{figure*}

%ch 1 We briefly recall the governing equations for modeling the elastic behavior of solids under loads. The reader is referred to standard textbooks  \cite{gurtin_fried_anand_2010} for a detailed discussion on the mechanics of continuous media.
\section{Governing Equations for Elasticity}
\label{sec:theory}
The governing equations for elasticity problems, in the absence of inertial forces, are  given as follows:
\begin{align}
	\begin{split}
		Div{\bfsigma} & + \bm{B} = \bm{0}, ~~\text{in}~~\Omega,\\
		\bfsigma =  \bbC : \bfepsilon, ~~&~~ \bfepsilon = \frac{1}{2} \left( \bfnabla \bfu + (\bfnabla \bfu)^T\right),\\
		\bfsigma \bfn = \bft_{bc} ~~ \text{on}~~ & \partial\Omega_{N} \text{~~and~~}	\bfu = \bfu_{bc}~~ \text{on}~~\partial\Omega_{D}.
	\end{split}
	\label{eq:sys1}
\end{align}
In the above, $\bfsigma$ and $\bfepsilon$ denotes the stress and the (linearized) strain in the material.  $\bfu$ and $\bfB$ denotes the displacement vector and body force vector (per unit volume), respectively. $\Omega$ denotes the volumetric domain, $Div$ denotes the divergence operator, and $\bbC$ is the fourth order elasticity tensor.  $\bft_{bc}$ and $\bfu_{bc}$ denote the known traction and displacement vectors on the (non-overlapping) parts of the boundary $\partial\Omega_{N}$ and $\partial\Omega_{D}$, respectively. $\bfn$ denotes the unit outward normal on the external boundary $\partial\Omega$.  Under two-dimensional plane-strain conditions, the  unknown components for displacement vector and stress tensor are $(u_x, u_y)$ and $(\sigma_{xx}$, $\sigma_{yy}$, $\sigma_{xy})$, respectively.

%The total strain $\bfepsilon$ is calculated from displacement gradients  as follows $ \bfepsilon = 0.5(\bfnabla\bfu + (\bfnabla\bfu)^T)$. Assuming small deformation, $\bfepsilon$ is  decomposed into the sum of elastic and plastic strain  components denoted by $\bfepsilon^e$ and $\bfepsilon^p$ , respectively: $\bfepsilon = 	\bfepsilon^e + 	\bfepsilon^p$. The Cauchy stress is derived from the elastic strain tensor as $\bfsigma = \bbC:\bfepsilon^e$, where 

%\section{NN }
%\label{sec:nn}

%\textcolor{red}{ talk about two dimensions and what components}
%components $\bfu = (\bfu_1, \bfu_2)$
%components $\bfepsilon^p : (\epsilon^p_{xx}, \epsilon^p_{yy}, \epsilon^p_{xy}, \epsilon^p_{zz}
%$\bfsigma : (\sigma_{xx}, \sigma_{yy}, \sigma_{xy}, \sigma_{zz})$
%Due to the symmetry of the stress and (plastic) strain tensors,

%Moreover, since these quantities are related through the constitutive relationship, few potential architectures that can be employed to predict these fields in the domain include 1. $(\bfu, \bfsigma)$, 2. $(\bfu, \bfepsilon) $ or 3. $(\bfu, \bfsigma, \bfepsilon)$ as the outputs of the neural network. As shown later in Section \ref{sec:u_ps_error}, the choice of the architectures (i) and (ii)  does not lead to accurate prediction of the plastic strain and stress fields in the domain, respectively.  

\section{Model Architecture}
\label{sec:arch}
%We develop and train a physics-informed DL framework to approximate this mapping $\Psi$ such that $\mathcal{I}^{HR} \approx  \Psi(\mathcal{I}^{LR}; {\bfalpha}).$
%\begin{equation}
%	\mathcal{I}^{HR} \approx  \Psi(\mathcal{I}^{LR}; {\bfalpha}).
%\end{equation} 
%In the above, $\bfalpha$ represents the the vector of physical quantities such as material properties, boundary conditions, or geometry.  
We train physics-informed DL framework to approximate the mapping $\Psi : \mathcal{I}^{LR} \rightarrow \mathcal{I}^{HR}$ to reconstruct the HR deformation field ($\mathcal{I}^{HR}$) from the LR ($\mathcal{I}^{LR}$) data.  The two architectures evaluated in this study are i) Residual Dense Network (RDN) \cite{zhang2018residual}, and ii) FSRCNN \cite{dong2016accelerating}.  In this work, we use the following hyper-parameters for the RDN model: number of residual blocks: $2$, number of layers in each residual block: $4$, growth rate: $32$, and  number of features: $32$.  For the FSRCNN model, we use the following hyper-parameters: number of layers: $8$, and LR feature dimensions $d=128$ and $s=64$.  These hyper-parameters also ensure that both the models have (almost) same number of trainable parameters.  The inputs to both the models consist of LR data $\{u_x, u_y, \sigma_{xx}, \sigma_{yy}, \sigma_{xy}\}$ obtained by running simulations on a coarse mesh (see Fig.~\ref{fig:fig_army}) and then evaluating the solution (using underlying interpolating basis functions) on a $32\times32$ structured grid. The outputs of these models  correspond to the HR data on a $128\times128$ structured grid as shown in Figure \ref{fig:fig_army}.

\section{Constructing the Loss Function}
\label{sec:loss}
For the unsupervised model, wherein the HR labeled data is not needed, the total network loss $\mathcal{L}$ is obtained only from the physics-based constraints corresponding to the governing equations and boundary conditions. For the mixed-variable formulation (displacement vector $\bfu$ and stress tensor $\bfsigma$ as outputs), the total loss function $\mathcal{L}$ is constructed as follows 
\begin{align}
	\begin{split}
		\mathcal{L} = &\lambda_1\, \underbrace{|| \bfnabla\cdot\bfsigma ||}_{\text{PDE}} \, + \,  \lambda_2 \,\underbrace{|| \bfsigma-\bbC:\bfepsilon ||}_{\text{Constitutive law}} \\ &+  \, \lambda_3\, \underbrace{|| \bfu - \bfu_{bc}||_{\partial\Omega_{U}}}_{\text{Dirichlet BC}} + \, \lambda_4\, \underbrace{|| \bfsigma\bfn - \bft_{bc} ||_{\partial\Omega_{N}}}_{\text{Neumann BC}},
	\end{split}
\end{align}
where $||(\cdot)||$ denotes the $L^1$ norm of the quantity $(\cdot)$.  $L^1$ norm is chosen to make the model robust to noise and outliers in the LR data. The scalar constants $\lambda_1$, $\lambda_2$, $\lambda_3$, and $\lambda_4$ are  chosen to nondimensionalize the individual loss components. In this work, we choose $\lambda_1 = \frac{H}{\mu}$,~$\lambda_2 = \frac{1}{\mu}$, $\lambda_3 = \frac{20}{ U_0}$, and $\lambda_4 = \frac{20}{\mu}$, where $\mu$ and $U_0$ represent the shear modulus and the characteristic displacement in the body, respectively. $H$ denotes the height of the body. Relatively larger magnitudes of $\lambda_3$ and $\lambda_4$ are chosen to assign more weight to the boundary conditions. 

Two neural networks based on RDN and FSRCNN architectures are implemented and trained using PyTorch framework \cite{NEURIPS2019_9015}.  The network's total loss $\mathcal{L}$ is minimized by iteratively updating trainable parameters by using Adam optimizer \cite{kingma_adam:_2015} for around $2000$ epochs with learning rate $\eta = 10^{-4}$. We also use \texttt{ReduceLROnPlateau} scheduler with the \texttt{patience}  $=30$. L-BFGS algorithm \cite{zhu1997algorithm} is then used for local fine-tuning of the solution until loss converges.  The training is performed using  NVIDIA Quadro RTX 8000 graphics card and takes around $8$ and $14$ hours for RDN and FSRCNN models, respectively. The source code and the dataset used in this research can be found at \url{https://github.com/sairajat/SR_LE} ~upon acceptance of this paper.

%Limited-memory Broyden-Fletcher-Goldfarb-Shanno 
% ( 

%\textcolor{red}{We note that the spatial variation of the deformation field is non-linear and the input data has large error}

% \begin{figure}[H]
	% 	\centering
	% 	\begin{subfigure}[b]{.495\linewidth}
		% 		\centering
		% 		{\includegraphics[width=.50\linewidth]{./figures/coarse_mesh.pdf}}
		% 		\caption{Coarse mesh.}
		% 	\end{subfigure}%
	% 	\begin{subfigure}[b]{.495\linewidth}
		% 		\centering
		% 		{\includegraphics[width=.50\linewidth]{./figures/fine_mesh.pdf}}
		% 		\caption{Fine mesh.}
		% 	\end{subfigure}\\
	% 	\caption{Schematic showing the geometry and the applied boundary conditions.}
	% \end{figure}

%\subsection{Model validation}

\begin{figure*}[t]
	\centering
	{\includegraphics[width=.95\linewidth]{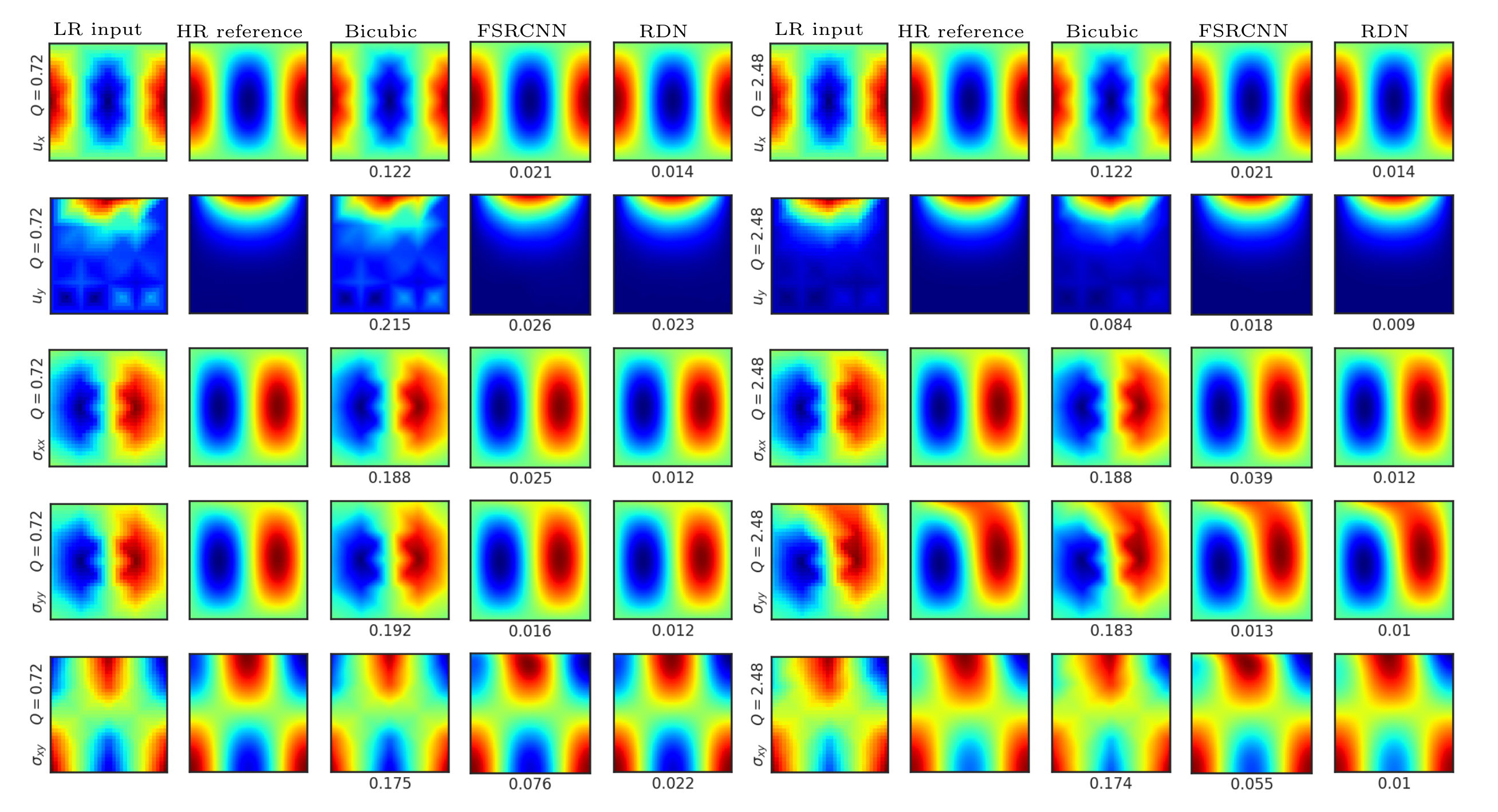}}
		\caption{The color contours of displacement vector and stress tensor components in two-dimensional elastic deformation 
		reconstructed with physics-informed super-resolution frameworks. Values below the plots indicate the $L_2$ error  $e$. In both the blocks, the LR input data, HR ground truth data, bicubic interpolation, FSRCNN output, and the RDN output are plotted from the left to the right.}
	\label{fig:results}
\end{figure*}

\section{Results \& Discussion}
\label{sec:results}
In what follows, we demonstrate the effectiveness of SR framework in reconstructing HR displacement and stress fields from LR input data for linear elastic simulations -- which we believe is a first step  in demonstrating the strength
of machine-learned super-resolution techniques in solid mechanics.

%The performance of the proposed SR framework is evaluated by applying it to resolve
We apply the framework to resolve the stress and displacement fields within an isotropic body undergoing linear elastic deformation. The schematic of the body along with the boundary conditions is shown in Figure \ref{fig:fig_army}.  The body is assumed to deform quasi-statically under plane strain conditions with the body force vector $\bfB$ given as 
\begin{align*}
	B_x &= \lambda \left[ 4\pi^2 \cos(2\pi x) \sin(\pi y) - \pi \cos(\pi x) Q y^3 \right]  \\
	& \quad + \mu \left[ 9\pi^2 \cos(2\pi x) \sin(\pi y) - \pi \cos(\pi x) Q y^3 \right],\\
	B_y &= \lambda \left[ 2\pi^2 \sin(2\pi x) \cos(\pi y) - 3 \sin(\pi x) Q y^2 \right]\\
	& \quad + \mu \left[ -6 \sin(\pi x) Q y^2 + 2\pi^2 \sin(2\pi x) \cos(\pi y) \right. \\
	& \left. \quad + \, 0.25 \pi^2 \sin(\pi x) Q y^4 \right].
\end{align*}
In this work, the material constants $\lambda$ and $\mu$ are taken to be $1$ and $0.5$,  respectively. The quantity $Q \in [0, 4]$ affects the  boundary conditions (see Fig.~\ref{fig:fig_army}) and the body force $\bfB$. The characteristic displacement $U_0$ is taken to $1$ (maximum value of $u_y$ on the top boundary). The ground truth data is generated by solving the system of equations \eqref{eq:sys1} on a coarse mesh (shown in Fig.~\ref{fig:fig_army}) using Finite Element Method for $Q$ regularly sampled at an interval of $0.04$. The data is then randomly split in a $80:20$ ratio for training and test purposes. 

The framework super-resolves the deformation fields onto the $128\times128$ mesh, shown in  Fig.~\ref{fig:fig_army}, which is $\approx \!400$ times finer than the coarse mesh used to obtain the LR data. The HR outputs for each model are obtained by doing a forward propagation through the corresponding trained models. For comparison, along with HR labeled data, we also utilize a simple bicubic interpolation of fields. We note that the HR labeled data is used only for the comparison with the predicted outputs. 

Figure \ref{fig:results} presents the results for the reconstructed displacement and stress fields for $2$ different values of $Q$ for both the models. We can see that the both the frameworks are able to super-resolve all the deformation fields with great accuracy as the plots show great agreement with the reference HR ground truth data. To qualitatively measure the accuracy, we define a relative error measure as $e = \frac{||\mathcal{I}^{HR} - \Psi(\mathcal{I}^{LR})||_{L^2}}{||\mathcal{I}^{HR}||_{L^2}}$, where $\Psi(\mathcal{I}^{LR})$ is the model output. The values of $e$ are reported underneath the reconstructed fields obtained using the SR frameworks and the bicubic interpolation. As can be seen, the error is largest for the bicubic interpolated data as compared to both the physics-informed models.  This is expected since the interpolated data may not faithfully satisfy the  governing laws of the system.  We also notice that the error is larger for FSRCNN based model as compared to RDN based model. The reconstructed HR outputs obtained from the RDN based model  almost match the accuracy of an advanced numerical solver running at  $400$ times the coarse mesh resolution. We believe that the better accuracy for the RDN model results from the use of residual connections and smaller kernel sizes during convolution and upsampling operations in its architecture.  This validates the concept that a deep-learning based physics-informed SR framework can be used to faithfully reconstruct the fields at a higher resolution while simultaneously satisfying the governing laws.  We note that the proposed physics-informed SR strategy can be easily extended to non-rectangular domains \cite{gao2020phygeonet} or account for boundary conditions in a \textit{hard} manner \cite{rao2021physics}.

%Therefore, we successfully trained and evaluated two deep learning based physics-informed super-resolution frameworks to reconstruct high-resolution stress and displacement fields from low-resolution simulation data with large discretization errors. 

\section{Conclusion}
\label{sec:conclusion}
In summary, we successfully trained and evaluated two physics-informed super-resolution frameworks based on Residual Dense Network  \cite{zhang2018residual} and FSRCNN \cite{dong2016accelerating} architectures to super-resolve the deformation fields in a body undergoing elastic deformation.  Among the two deep learning architectures evaluated in this work, we show that the framework based on RDN is more accurate and matches the accuracy of an advanced numerical solver running at  $400$ times the coarse mesh resolution (see Figs.~\ref{fig:fig_army} and \ref{fig:results}).  The approach is successfully able to learn high-resolution spatial variation of displacement and stress fields from their low-resolution counterparts for the linear elastic case discussed.  These advantages are possible due to the combined effect of two rapidly evolving research areas - Physics informed neural networks \cite{raissi2017physicsI, raissi2019physics, arora2022physics} and computer vision \cite{voulodimos2018deep}.  We emphasize that the current work focuses on the demonstration of feasibility of the concept while the assessment of potential computational advantages, including the extension  to hyperelastic deformation, is deferred to future research.

The approach exemplifies how machine-learning can be leveraged to conduct such mechanical calculations for materials with complex constitutive response (eg. dislocation mediated plastic deformation and fracture modeling \cite{arora2020finite, nielsen2019finite, niordson2019homogenized, joshi2020equilibrium,  arora2020dislocation, yingjun2016phase, arora2019computational, kuroda2008finite, lynggaard2019finite, arora2020unification, borden2014higher, arora2022mechanics})  to reduce the computational complexity and accelerate scientific discovery and engineering design.

% Use \bibliography{yourbibfile} instead or the References section will not appear in your paper

\section*{Acknowledgments}
This work was conceptualized during the author's time at Carnegie Mellon University (CMU). The author thank Ankit Shrivastava, Ph.D. candidate at CMU, for useful discussions and comments on the manuscript. %, who is currently a Ph.D. candidate at Carnegie Mellon University, Pittsburgh.%, and will be joining Sandia National Labs as a Postdoctoral fellow in February 2022.

\bibliographystyle{alpha}
\bibliography{main}

\newcommand{\etalchar}[1]{$^{#1}$}
\begin{thebibliography}{PGM{\etalchar{+}}19}

\bibitem[AA20a]{arora2020dislocation}
Rajat Arora and Amit Acharya.
\newblock Dislocation pattern formation in finite deformation crystal
  plasticity.
\newblock {\em International Journal of Solids and Structures}, 184:114--135,
  2020.

\bibitem[AA20b]{arora2020unification}
Rajat Arora and Amit Acharya.
\newblock A unification of finite deformation {J2} {V}on-{M}ises plasticity and
  quantitative dislocation mechanics.
\newblock {\em Journal of the Mechanics and Physics of Solids}, 143:104050,
  2020.

\bibitem[AAA22]{arora2022mechanics}
Abhishek Arora, Rajat Arora, and Amit Acharya.
\newblock Mechanics of micropillar confined thin film plasticity.
\newblock {\em arXiv preprint arXiv:2202.06410}, 2022.

\bibitem[AKDC22]{arora2022physics}
Rajat Arora, Pratik Kakkar, Biswadip Dey, and Amit Chakraborty.
\newblock Physics-informed neural networks for modeling rate-and
  temperature-dependent plasticity.
\newblock {\em arXiv preprint arXiv:2201.08363}, 2022.

\bibitem[Aro19]{arora2019computational}
Rajat Arora.
\newblock {\em Computational Approximation of Mesoscale Field Dislocation
  Mechanics at Finite Deformation}.
\newblock PhD thesis, Carnegie Mellon University, 2019.

\bibitem[AZA20]{arora2020finite}
Rajat Arora, Xiaohan Zhang, and Amit Acharya.
\newblock Finite element approximation of finite deformation dislocation
  mechanics.
\newblock {\em Computer Methods in Applied Mechanics and Engineering},
  367:113076, 2020.

\bibitem[BGL{\etalchar{+}}19]{bode2019using}
Mathis Bode, Michael Gauding, Zeyu Lian, Dominik Denker, Marco Davidovic,
  Konstantin Kleinheinz, Jenia Jitsev, and Heinz Pitsch.
\newblock Using physics-informed super-resolution generative adversarial
  networks for subgrid modeling in turbulent reactive flows.
\newblock {\em arXiv preprint arXiv:1911.11380}, 2019.

\bibitem[BHLV14]{borden2014higher}
Michael~J Borden, Thomas~JR Hughes, Chad~M Landis, and Clemens~V Verhoosel.
\newblock A higher-order phase-field model for brittle fracture: Formulation
  and analysis within the isogeometric analysis framework.
\newblock {\em Computer Methods in Applied Mechanics and Engineering},
  273:100--118, 2014.

\bibitem[DHLK19]{deng2019super}
Zhiwen Deng, Chuangxin He, Yingzheng Liu, and Kyung~Chun Kim.
\newblock Super-resolution reconstruction of turbulent velocity fields using a
  generative adversarial network-based artificial intelligence framework.
\newblock {\em Physics of Fluids}, 31(12):125111, 2019.

\bibitem[DLT16]{dong2016accelerating}
Chao Dong, Chen~Change Loy, and Xiaoou Tang.
\newblock Accelerating the super-resolution convolutional neural network.
\newblock In {\em European conference on computer vision}, pages 391--407.
  Springer, 2016.

\bibitem[EAK{\etalchar{+}}20]{esmaeilzadeh2020meshfreeflownet}
Soheil Esmaeilzadeh, Kamyar Azizzadenesheli, Karthik Kashinath, Mustafa
  Mustafa, Hamdi~A Tchelepi, Philip Marcus, Mr~Prabhat, Anima Anandkumar,
  et~al.
\newblock Meshfreeflownet: a physics-constrained deep continuous space-time
  super-resolution framework.
\newblock In {\em SC20: International Conference for High Performance
  Computing, Networking, Storage and Analysis}, pages 1--15. IEEE, 2020.

\bibitem[FFT19]{fukami2019super}
Kai Fukami, Koji Fukagata, and Kunihiko Taira.
\newblock Super-resolution analysis with machine learning for low-resolution
  flow data.
\newblock In {\em 11th International Symposium on Turbulence and Shear Flow
  Phenomena, TSFP 2019}, 2019.

\bibitem[FFT21]{fukami2021machine}
Kai Fukami, Koji Fukagata, and Kunihiko Taira.
\newblock Machine-learning-based spatio-temporal super resolution
  reconstruction of turbulent flows.
\newblock {\em Journal of Fluid Mechanics}, 909, 2021.

\bibitem[GSW20]{gao2020phygeonet}
Han Gao, Luning Sun, and Jian-Xun Wang.
\newblock Phygeonet: Physics-informed geometry-adaptive convolutional neural
  networks for solving parametric pdes on irregular domain.
\newblock {\em arXiv e-prints}, pages arXiv--2004, 2020.

\bibitem[GSW21]{gao2021super}
Han Gao, Luning Sun, and Jian-Xun Wang.
\newblock Super-resolution and denoising of fluid flow using physics-informed
  convolutional neural networks without high-resolution labels.
\newblock {\em Physics of Fluids}, 33(7):073603, 2021.

\bibitem[JABG20]{joshi2020equilibrium}
Tushar Joshi, Rajat Arora, Anup Basak, and Anurag Gupta.
\newblock Equilibrium shape of misfitting precipitates with anisotropic
  elasticity and anisotropic interfacial energy.
\newblock {\em Modelling and Simulation in Materials Science and Engineering},
  28(7):075009, 2020.

\bibitem[KB15]{kingma_adam:_2015}
Diederik~P. Kingma and Jimmy Ba.
\newblock Adam: {A} {Method} for {Stochastic} {Optimization}.
\newblock {\em ICLR}, 2015.

\bibitem[KT08]{kuroda2008finite}
M.~Kuroda and V.~Tvergaard.
\newblock A finite deformation theory of higher-order gradient crystal
  plasticity.
\newblock {\em Journal of the Mechanics and Physics of Solids},
  56(8):2573--2584, 2008.

\bibitem[LNN19]{lynggaard2019finite}
J.~Lynggaard, K.~L. Nielsen, and C.~F. Niordson.
\newblock Finite strain analysis of size effects in wedge indentation into a
  face-centered cubic (fcc) single crystal.
\newblock {\em European Journal of Mechanics / A Solids}, 76:193--207, 2019.

\bibitem[NN19]{nielsen2019finite}
K.~L. Nielsen and C.~F. Niordson.
\newblock A finite strain fe-implementation of the fleck-willis gradient
  theory: Rate-independent versus visco-plastic formulation.
\newblock {\em European Journal of Mechanics / A Solids}, 75:389--398, 2019.

\bibitem[NT19]{niordson2019homogenized}
C.~F. Niordson and V.~Tvergaard.
\newblock A homogenized model for size-effects in porous metals.
\newblock {\em Journal of the Mechanics and Physics of Solids}, 123:222--233,
  2019.

\bibitem[PGM{\etalchar{+}}19]{NEURIPS2019_9015}
Adam Paszke, Sam Gross, Francisco Massa, Adam Lerer, James Bradbury, Gregory
  Chanan, Trevor Killeen, Zeming Lin, Natalia Gimelshein, Luca Antiga, Alban
  Desmaison, Andreas Kopf, Edward Yang, Zachary DeVito, Martin Raison, Alykhan
  Tejani, Sasank Chilamkurthy, Benoit Steiner, Lu~Fang, Junjie Bai, and Soumith
  Chintala.
\newblock Pytorch: An imperative style, high-performance deep learning library.
\newblock In H.~Wallach, H.~Larochelle, A.~Beygelzimer, F.~d\textquotesingle
  Alch\'{e}-Buc, E.~Fox, and R.~Garnett, editors, {\em Advances in Neural
  Information Processing Systems 32}, pages 8024--8035. Curran Associates,
  Inc., 2019.

\bibitem[RPK17]{raissi2017physicsI}
Maziar Raissi, Paris Perdikaris, and George~Em Karniadakis.
\newblock Physics informed deep learning (part i): Data-driven solutions of
  nonlinear partial differential equations.
\newblock {\em arXiv preprint arXiv:1711.10561}, 2017.

\bibitem[RPK19]{raissi2019physics}
Maziar Raissi, Paris Perdikaris, and George~E Karniadakis.
\newblock Physics-informed neural networks: A deep learning framework for
  solving forward and inverse problems involving nonlinear partial differential
  equations.
\newblock {\em Journal of Computational Physics}, 378:686--707, 2019.

\bibitem[RSL21]{rao2021physics}
Chengping Rao, Hao Sun, and Yang Liu.
\newblock Physics-informed deep learning for computational elastodynamics
  without labeled data.
\newblock {\em Journal of Engineering Mechanics}, 147(8):04021043, 2021.

\bibitem[SW20]{sun2020physics}
Luning Sun and Jian-Xun Wang.
\newblock Physics-constrained bayesian neural network for fluid flow
  reconstruction with sparse and noisy data.
\newblock {\em Theoretical and Applied Mechanics Letters}, 10(3):161--169,
  2020.

\bibitem[SWB{\etalchar{+}}20]{subramaniam2020turbulence}
Akshay Subramaniam, Man~Long Wong, Raunak~D Borker, Sravya Nimmagadda, and
  Sanjiva~K Lele.
\newblock Turbulence enrichment using physics-informed generative adversarial
  networks.
\newblock {\em arXiv preprint arXiv:2003.01907}, 2020.

\bibitem[VDDP18]{voulodimos2018deep}
Athanasios Voulodimos, Nikolaos Doulamis, Anastasios Doulamis, and Eftychios
  Protopapadakis.
\newblock Deep learning for computer vision: A brief review.
\newblock {\em Computational intelligence and neuroscience}, 2018, 2018.

\bibitem[XFCT18]{xie2018tempogan}
You Xie, Erik Franz, Mengyu Chu, and Nils Thuerey.
\newblock tempogan: A temporally coherent, volumetric gan for super-resolution
  fluid flow.
\newblock {\em ACM Transactions on Graphics (TOG)}, 37(4):1--15, 2018.

\bibitem[YZL{\etalchar{+}}16]{yingjun2016phase}
Gao Yingjun, Luo Zhirong, Huang Lilin, Mao Hong, Huang Chuanggao, and Lin Kui.
\newblock Phase field crystal study of nano-crack growth and branch in
  materials.
\newblock {\em Modelling and Simulation in Materials Science and Engineering},
  24(5):055010, 2016.

\bibitem[ZBLN97]{zhu1997algorithm}
Ciyou Zhu, Richard~H Byrd, Peihuang Lu, and Jorge Nocedal.
\newblock Algorithm 778: L-bfgs-b: Fortran subroutines for large-scale
  bound-constrained optimization.
\newblock {\em ACM Transactions on mathematical software (TOMS)},
  23(4):550--560, 1997.

\bibitem[ZTK{\etalchar{+}}18]{zhang2018residual}
Yulun Zhang, Yapeng Tian, Yu~Kong, Bineng Zhong, and Yun Fu.
\newblock Residual dense network for image super-resolution.
\newblock In {\em Proceedings of the IEEE conference on computer vision and
  pattern recognition}, pages 2472--2481, 2018.

\end{thebibliography}

\end{document}